# Smart Pointers and Shared Memory Synchronisation for Efficient Inter-process Communication in ROS on an Autonomous Vehicle

Costin Iordache[1], Stephen M. Fendyke[1], Mike J. Jones[1], Robert A. Buckley[1]

*Abstract*—Despite the stringent requirements of a real-time system, the reliance of the Robot Operating System (ROS) on the loopback network interface imposes a considerable overhead on the transport of high bandwidth data, while the nodelet package, which is an efficient mechanism for intra-process communication, does not address the problem of efficient local inter-process communication (IPC). To remedy this, we propose a novel integration into ROS of smart pointers and synchronisation primitives stored in shared memory. These obey the same semantics and, more importantly, exhibit the same performance as their C++ standard library counterparts, making them preferable to other local IPC mechanisms. We present a series of benchmarks for our mechanism - which we call LOT (Low Overhead Transport) - and use them to assess its performance on realistic data loads based on Five's Autonomous Vehicle (AV) system, and extend our analysis to the case where multiple ROS nodes are running in Docker containers. We find that our mechanism performs up to two orders of magnitude better than the standard IPC via local loopback. Finally, we apply industry-standard profiling techniques to explore the hotspots of code running in both user and kernel space, comparing our implementation against alternatives.

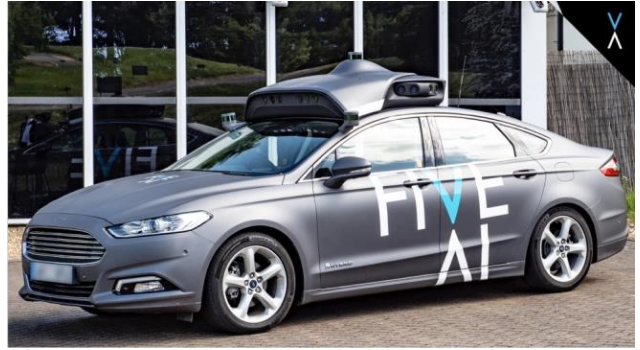

Fig. 1: Our prototype autonomous vehicle development platform includes cameras providing surround stereo vision capability, lidars, an industrial grade IMU, a GNSS localisation system and surround radar.

## I. INTRODUCTION

Complex robotic systems typically rely on at least two main layers of software: the operating system (OS), whose role is to expose an interface through which the underlying hardware is managed, and the middleware, which adds services and capabilities to meet requirements of the applications. Further layers of abstraction may exist in the form of, for instance, software being run in containers using Docker. As too many abstraction layers may hurt performance, the challenge for developers is to find the right balance between the amount of abstraction and the inherent overhead and to tune the software at all levels.

The requirement on real-time robotic systems to be deterministic in their behaviour [4] can only be honoured if the communication - both in terms of throughput and latency - between the system components is bounded and predictable. This is particularly true for robots with sensors generating large quantities of data or data at high frequencies.

Five's prototype autonomous vehicle (AV) shown in Fig. 1 is an example of such a system and provides the context and initial motivation for this work. Its sensor set includes cameras providing surround vision capability, multiple Lidars, an industrial grade IMU, a GNSS localisation system and surround radar. The AV software stack runs in a series of Docker containers on a high performance computer. Large quantities of data flow through this system, most of which comes from a relatively small number of high-bandwidth sensors operating at between 10Hz and 30Hz: principally the cameras (which can contribute up to 300MB/s each), but also Lidar and Radar (up to ~10 MB/s in aggregate). A number of lower bandwidth sensors operate at higher sample rates but are sensitive to latency, such as the GNSS receiver and IMU.

In this paper, we present a generally applicable approach to optimising the communication between ROS nodes deployed in separate OS processes on the same machine, such as those which together form our software stack capable of handling all the perception, planning and control systems of an autonomous vehicle (AV) in real-time. We demonstrate that the overhead of local inter-process communication can be considerably reduced by combining Linux shared memory, C++ smart pointers, and synchronisation primitives such as mutexes and thread-safe queues, and propose a serialization-free, zero-copy local IPC mechanism, that is both predictable and reliable with much smaller communication overhead than existing solutions.

As communication between ROS nodes depends on both the OS and the underlying hardware, it is paramount to consider the strengths and weaknesses of the software layers alongside those of the hardware architecture. We describe the background to this work in section II, in which we discuss overheads first in Linux and then in ROS. Throughout this section, we review relevant related work. We describe our LOT mechanism in section III and present an analysis in section IV. Finally, section V concludes the paper and offers suggestions for future work.

[1] All authors are with Five AI Ltd, Bristol & Exeter House, Bristol BS1 6QS, UK. Corresponding author: costin.iordache@five.ai

## II. BACKGROUND AND RELATED WORK

### A. Overheads in Linux

The Linux kernel tracks the hardware and software resources used by each user process, including their address space and threads. The address space is a contiguous virtual memory space assigned by the kernel to each user process, and is divided into memory areas which can be dynamically added and removed. Programs are loaded from the storage medium into this address space and executed in user-space. While user-space code cannot invade the kernel space or other user processes, special system calls can be invoked to request the kernel to act on behalf of the user process. Each kernel-user space context switch, of course, requires extra CPU cycles.

Threads represent the smallest runnable unit that the kernel can schedule for execution and an abstraction that the programmer can use to enable genuine parallelism on multiprocessor machines. In Linux, threads are implemented as processes sharing a common address space with other sibling processes [18, 19]. This has the consequence that context switches between threads belonging to different processes incur extra cost because, in addition to swapping the processor state, they require the process' virtual address space to be swapped too.

Communication between threads may thus be intra-process or inter-process. Intra-process communication has less overhead, being coordinated with highly efficient primitives such as mutexes, semaphores and atomics. For the inter-process case, several options exist. If the processes reside on different machines then the kernel, the network stack and network interface controller (NIC) must all be involved in the data communication process. However, when processes are hosted on the same machine they can communicate locally through the network loopback interface, Unix domain sockets (UDS) or shared memory, as well as by using synchronisation primitives, pipes, files and signals, though not all these mechanisms are conceptually suitable for real-time robotics systems or give good performance. In the case of the loopback interface mechanism, for instance, the communication overhead becomes significant for data blocks larger than 64K bytes as reported by Maruyama et al. [1]. To understand this, we consider the architecture of the network stack and briefly explore the path of data flow.

Conceptually the TCP/IP protocol suite comprises application, transport, internet, link and physical layers [14, 15]. The transport layer exposes TCP and UDP sockets, which act as endpoints for the transmission of data across the network to processes which may be hosted either by the same or a different machine. We note that data transmission requires copying data from user to kernel space. From there data is passed to the link layer which further copies it to the physical layer by means of NIC driver calls. At the receiver, the reverse operation takes place resulting in data being copied from the NIC into kernel space, and then from kernel to user space. Finally, at the application layer, data is copied one more time into the user's buffer by the *tcp_recvmsg()* system call.

Before structured data can reach the transport layer to be transmitted, either locally or remotely, it needs to be flattened into a string of bytes at the sending end, a process known as serialisation. At the receiving end the inverse process takes place (deserialisation). These operations that are often functionally equivalent to a data copy may occur multiple times if there is more than one subscriber. For example, under the above model, three copies of an 8MB *Image* may be transmitted to the perception, object detection and persistence modules of an AV stack.

In addition to the overhead incurred by copying data multiple times, further performance degradation can be caused by data fragmentation at the IP layer. The largest size of IP datagram which can be transferred in one transaction is defined by the maximum transmission unit (MTU), which is typically up to 64KB [10, 11] for the loopback interface, and 1500 bytes for other interfaces. Packets larger than the MTU are fragmented into multiple datagrams for transmission and reassembled at the receiver.

These three processes - (de)serialisation, data copies between user and kernel space and data fragmentation at the IP layer - are the major sources of performance overhead as data is moved through the network layers. Additional overhead is then contributed by the physical layer when data is transmitted between processes hosted by different machines.

Efficient local data transport is also possible without using the network stack, by leveraging other features of the OS. In their performance evaluation of pipes, local sockets and shared memory, for instance, Venkataraman et al. [16] conclude that a shared memory mechanism is by far the most effective, both in terms of latency and throughput, except for data payloads smaller than 64KB where they found performance to be comparable. Unix domain sockets (UDS) facilitate local IPC by means of the OS kernel without using the network stack and all the associated overhead that comes with it: ACKs, encapsulation, flow control, routing, MTU and context switches in network interrupt service routines [17]. We note that these overheads are incurred regardless of the type of network interface being used, loopback or remote. Zhang et al. [27] and Redis [26] report a significant throughput improvement when switching from TCP/IP loopback to UDS.

### B. Overheads in ROS

The ROS framework [6] design encourages software to be split into components, or ROS *nodes*, each in its own OS process. Each ROS node communicates by publishing or subscribing to any number of strongly typed, unidirectional messaging channels known as *topics*. ROS is well-suited for use in distributed systems with ROS nodes on multiple compute nodes communicating over a network. Such systems are attractive when operating in constrained environments, providing good CPU performance where it is impractical to deploy a single powerful computer for reasons of space, power, or manufacturing cost. However, where such constraints do not exist, all the ROS nodes in a system may be run on a single machine. In this case, it is reasonable to consider whether the overhead of moving data between nodes could not be significantly smaller than in a distributed system given the absence of the network as the facilitator of data transmission. The ROS nodelet package [5] was introduced to improve performance in this scenario. It allows several ROS nodes to be run within a single OS process. However, this strategy still leaves some unsolved problems in that the implementation of the nodelet callback queue is not fully optimised [24], and the use of multiple processes may be required for other reasons, as in the case we discuss below. The use of ROS nodelets for efficient intra-process communication via shared pointers is also not possible in any robotic system that makes use of Docker[2] for containerisation, running system components in separate containers.

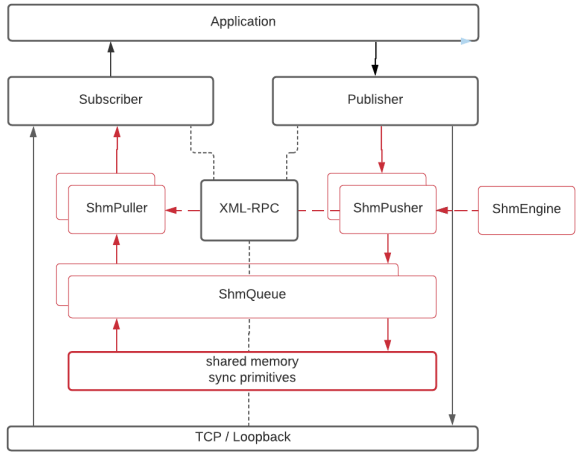

Fig. 2: LOTROS architecture

Fujita [12] extends ROS, adding a new topic transport protocol UDSROS which uses UDS (Unix domain sockets, discussed in section III) in place of TCP loopback. Although UDS shows some improvements over TCP sockets in terms of latency and throughput, the implementation still incurs overheads from message (de)serialisation and from copies between user and kernel space.

Wang et al. [2] propose a hybrid approach which they term TZC where large messages are split into two parts. A lightweight descriptor is transmitted over a ROS topic in the usual way, via TCPROS, while the main message payload is placed in shared memory. TZC's bespoke double reference counted scheme relies on a double-linked list stored in shared memory with reference counted nodes, and on *Boost*'s *shared_ptr* to deliver the payload in the standard ROS way. However, the reliance of TZC on TCP makes the approach sensitive to the connection establishment order and is the cause of unnecessary overhead compared to a standard solution [3]. We also note that the lack of a reliable abstraction to manage the lifetime of the message could cause a payload to dangle, resulting in a memory leak if its descriptor is not delivered.

Many robotic systems have relied since their inception on ROS. For such use cases, this paper proposes a serialisation free, zero copy local IPC mechanism via shared pointers stored in memory segments shared by several ROS nodes, that is one or two orders of magnitude faster than the standard ROS mechanisms.

### III. LOW OVERHEAD TRANSPORT (LOT)

In this section, we introduce our LOT mechanism, which we show schematically in Fig. 2.

#### A. C++ Smart Pointers

The C++ standard library (STL) includes smart pointers which "enable automatic, exception-safe, object lifetime management" [21] with different types of ownership semantics. In this paper, we are interested in the exclusive (*unique_ptr*) and shared (*shared_ptr*) ownership semantics. The *unique_ptr*'s performance and size are equivalent to those of raw pointers, as described by Meyers [20]. The *shared_ptr* encapsulates two raw pointers to keep track of the resource and a *control block*. The control block is by default allocated in the heap alongside the owned resource, and holds housekeeping data such as the reference counter, weak counter and (de)allocator [20]. Its size is typically a few bytes and because it is allocated in tandem with the managed object, the added overhead required by the extra allocation is close to zero. Similarly, because atomic updates of the reference counter are delegated to the hardware, the incurred runtime overhead is negligible. The destruction of the managed resource is accomplished via a virtual method whose dynamic dispatch overhead is also tiny. We conclude that these modest costs of smart pointers pale in comparison with the benefits, i.e. automatic lifetime management of dynamically allocated resources, preventing memory leaks.

#### B. Placing C++ Objects in Shared Memory

Sharing memory between several processes requires mapping a memory segment from the OS into the address space of each process requiring access. As the memory segment gets mapped to a different virtual address range in each process, subtle restrictions on C++ objects storing pointers to other objects arise: the pointed-to objects must be available in the same memory segment as their parent, while processes consuming a pointer must map it into their own address space before the first use. This makes raw pointers and STL smart pointers unsuitable for shared memory. To address this shortcoming, we use the *offset_ptr* from the industry-standard *Boost.Interprocess* library which refines the semantics such that pointers stored in objects placed in shared memory denote a relative offset with respect to their parent object rather than an absolute address in the process address space [22]. This additionally avoids impractical and error-prone manipulation of raw pointers in shared memory. Whilst *offset_ptr* allows the user to place pointers in shared memory, we note that it cannot be used for objects with virtual (polymorphic) functions, since the virtual dispatch mechanism used to implement them is unaware of the *offset_ptr*, being fixed in the C++ compiler implementation. Although this limitation can be overcome [29, 30], no such modification is required in our mechanism, as our ROS messages are not dependent on virtual methods or other compiler specific immutable features.

---

[2] Docker is a containerisation platform in which each container hosts one or more processes, and is run from an image built and tested ahead of time, which contains all required runtime dependencies. This provides benefits in terms of software development and reliable deployment.

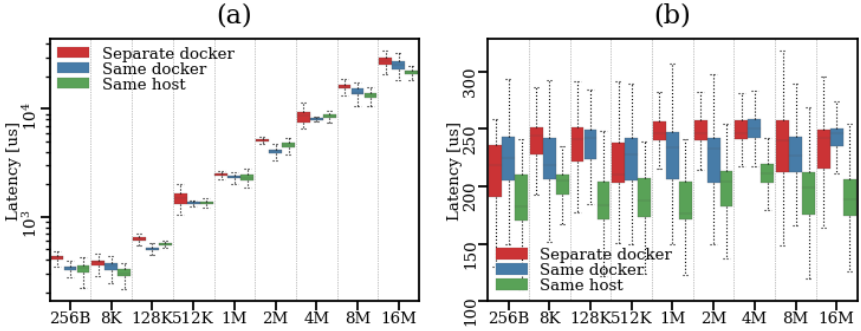
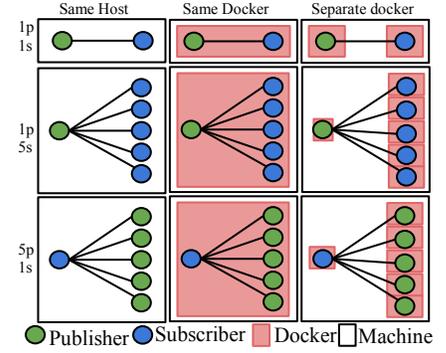

Fig. 3: Here we show the difference in latency between benchmark tests run natively on the same host, run on the same host within the same Docker container and on the same host within separate Docker containers for the cases of (a) TCP transport and (b) our LOT transport.

Fig. 4: Our execution environments consist of OS, ROS and Docker software abstractions which are combined as shown.

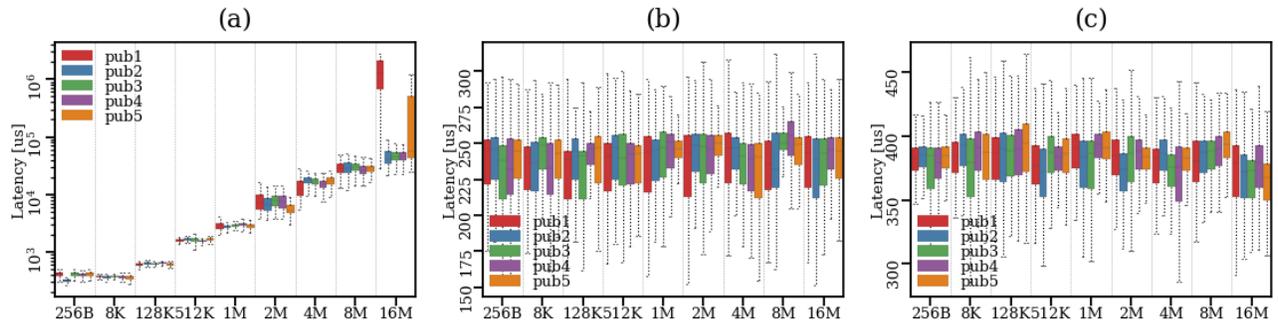

Fig. 5: Here we show the latency in the case of five publishers and a single subscriber (*5p1s*) in the cases of (a) TCP transport, (b) our LOT transport and (c) TZC transport. In all cases, we enforce connection order. Note that in the TCP case, latency is sensitive to the image size with publishers being treated fairly except in the 16MB payload case. For LOT and TZC, the latency is not sensitive to the image size and the gap between successive publishers is around 40us, but the LOT average latency is 235us whilst that of TZC is 385us.

### C. Integrating shared memory with ROS

In addition to using *Boost.Interprocess* to make ROS messages shared memory compliant as described in the previous section, inspired by Williams [23] we combine shared memory compliant mutexes and condition variables to create a generic, shared memory compliant queue - *ShmQueue* - the purpose of which is to provide thread safe access to the data it owns from multiple processes through a standard interface with well defined behaviour. In this context, thread safeness refers to the fact that multiple threads - which may be in different processes - can access the data owned by instances of *ShmQueue* such that the following conditions are met concomitantly: (a) each thread sees a consistent view of the data, (b) no data is lost or corrupted, (c) no race conditions arise and (d) threads can perform the same or distinct operations independently on the same or different instances of the queue.

To accommodate communication through shared memory, we have extended ROS's transport protocols by adding our LOT mechanism as LOTROS, under which ROS nodes rely on the XMLRPC [24] protocol and OS sockets to establish connections, and *ShmEngine* to transmit data. Each ROS node incorporates one *ShmEngine*. The LOTROS protocol settings, e.g. the names of *ShmQueue*s, are exchanged via XMLRPC during the connection setup. All ROS nodes using LOTROS share a common memory segment with configurable size and identified by a unique, well known string in the OS file system; e.g. */dev/shm/LOTROS*. When topics are advertised by the publisher, the *ShmEngine* assigns one *ShmPusher* per topic. When a node subscribes to a topic, a *ShmQueue* is created by the corresponding topic's *ShmPusher* and its name is returned to the subscriber. The subscriber, in turn, creates a *ShmPuller* per publisher and topic the main job of which is to spawn a thread, attach itself to the *ShmQueue* indicated by the publisher and start waiting for incoming items on the queue. Items may be of any shared memory compliant C++ type. These include the *ShmSharedPtr* and *ShmUniquePtr*, two of the shared memory compliant smart pointers we have defined. The *ShmPuller* also creates local private topics in order to forward items popped from the *ShmQueue* to the ROS subscriber using its standard interface. Unlike regular topics, the scope of private topics is limited to the node that created them, i.e. intra-process, and they are not publicly advertised. *ShmPuller* combines private topics and ROS intra-process message passing [24] by wrapping incoming items in *shared_ptr*s and publishing them as though they originated from another ROS node.

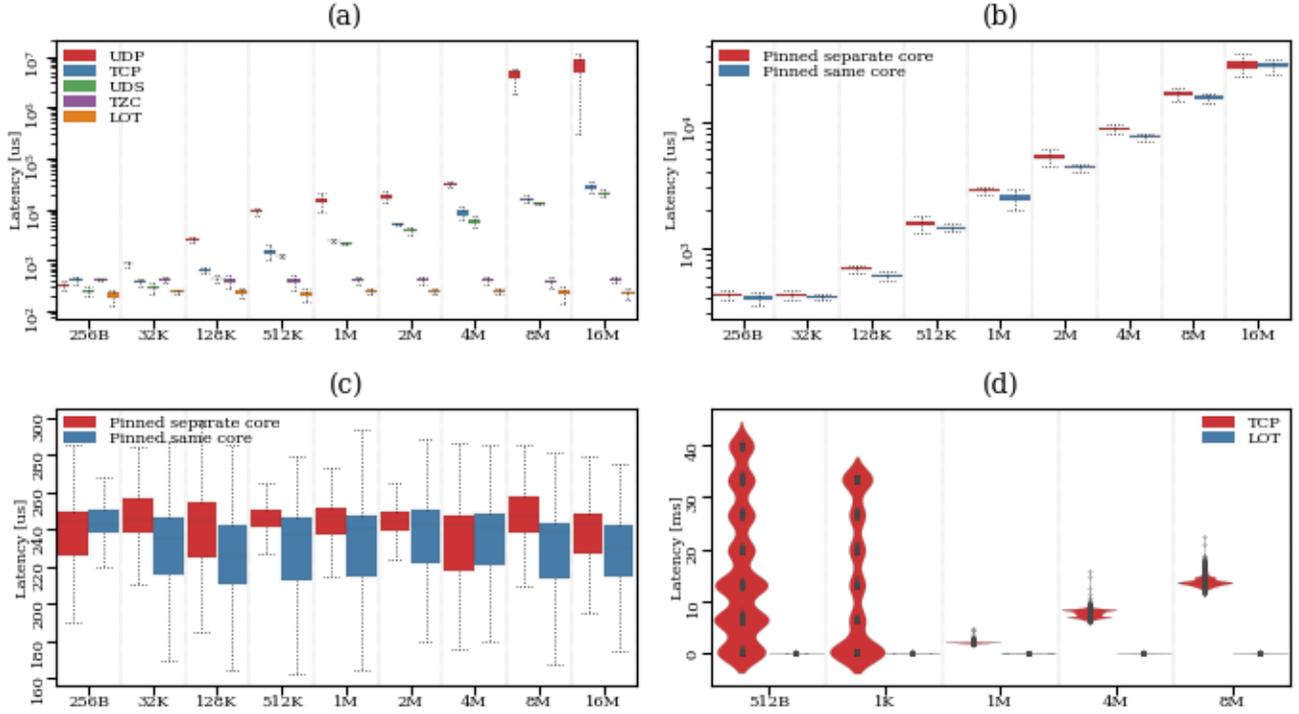

Fig. 6: We show the latency in the case of one publisher with one subscriber (*1p1s*) for various payload sizes. In (a) we present the UDP, TCP, UDS, LOT and TZC transports. In (b) we show the effect of pinning to the same vs different processors for TCP transport and the same in (c) for our LOT transport; note that pinning to the same processor core results in lower latency for TCP, especially for data larger than 32KB, only a tiny effect is observed in the LOT case, which also shows a negligible latency difference between small and large data. Finally in (d) we show the behaviour of TCP in buffering small data sending it in bursts and demonstrate the absence of this behaviour in the LOT case.

We note that the type of memory an object is allocated in (shared memory or otherwise) is irrelevant to that object's layout in memory, with the consequence that object serialisation - for instance, when message traffic on a ROS topic is saved to disk in a so-called *ROS bag* - is not impacted by our LOT data transport mechanism.

## IV. EVALUATION

In this section we present the robustness, latency and determinism characteristics of our LOT mechanism. We describe the results from a set of experiments inspired by our AV stack introduced in section I. Fig. 4 presents the execution environments that have been considered, with ROS nodes running in either *same* or *separate* Docker containers, or directly in the *same host OS*. The execution environments have been deployed on the computer described in section II. Within these environments we considered ROS graphs with a single publisher with a single subscriber (*1p1s*), a single publisher with five subscribers (*1p5s*) and five publishers with a single subscriber (*5p1s*). The ROS ecosystem does not allow the graph nodes to be started up in a deterministic order [9], so we built an ad-hoc system to enforce connection order establishment with the purpose of making the experiment results comparable. The end-to-end latencies are evaluated by transferring either *sensor_msgs::Image*s or *lot_msgs::Image*s with payload sizes ranging from 128B to 16MB. Each experiment consists of broadcasting 2000 *Image*s of fixed size at 30Hz and is discussed in the following subsections. Unless otherwise specified, ROS nodes have been executed in separate Docker containers. For clarity in Figs. 3, 5, 6 and 7 we do not show payload sizes of 512B, 1KB, 2Kb, 4KB and 8KB for which results are consistent with those shown. Furthermore, we refer to TCP/IP local IPC simply as TCP unless otherwise specified.

### A. UDP vs TCP vs UDS vs TZC vs LOT

Fig. 6(a) shows the distribution of latencies measured for a graph with one publisher and one subscriber (*1p1s*). For TCP and UDS transport mechanisms we see similar latency trends with values ranging from 200us to 26ms. For payload sizes up to 8KB the protocols exhibit a consistent performance pattern - LOT, UDS, UDP, TCP, TZC - with LOT having the smallest latency. The largest and smallest gap between median latencies of consecutive protocols is around 100us and 5us, respectively. For larger payloads, the protocol ranking changes to LOT, TZC, UDS, TCP, UDP and the latencies of TCP and UDS increase proportionally with the size of the

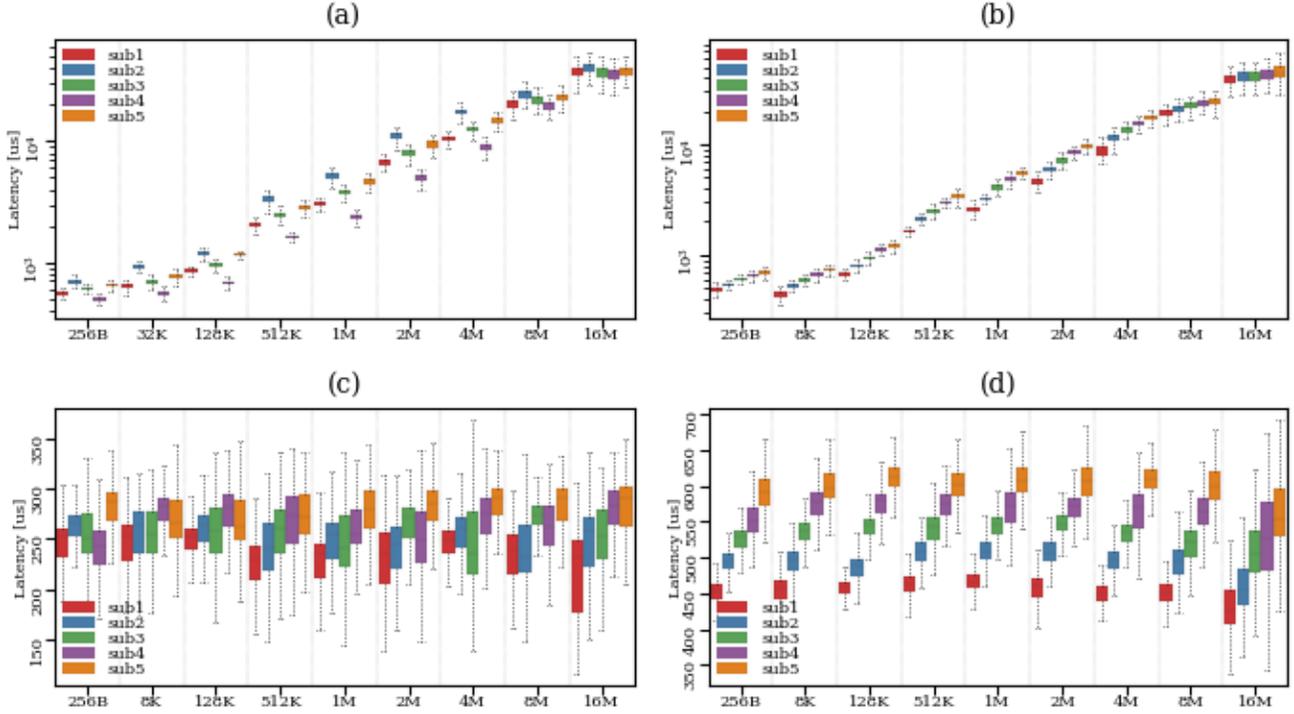

Fig. 7: We show the latency in the case of one publisher with five subscribers (*1p5s*) for various payload sizes in the case of (a) TCP transport with an undetermined connection order; (b) TCP transport, (c) our LOT transport and (d) TZC transport with enforced connection order. We note that while late TCP joiners incur higher latency of up to 7ms for 16MB, this latency is only 60us with LOT and 200us with TZC.

*Image*. This trend is also followed by UDP with the maximum being attained for 8MB and 16MB *Image*s where latencies of up to 6 seconds have been observed. Furthermore, because UDP provides no guarantees for message delivery, fewer than 2,000 *Image*s reached the subscriber. By contrast, LOT outperforms the other protocols except TZC by a few orders of magnitude, both in terms of the median latency, which is approximately 235us, and latency variance with *Image* size. In particular, LOT outperforms UDS, TCP and UDP by one order of magnitude for a 1MB payload, increasing to around two orders of magnitude for 16MB payloads, and TZC by approximately 57% - which corresponds to 170us - at all payload sizes.

*B. Late Joiners incur Higher Latency*

The boxplots in Fig. 7 show the latencies incurred in a *1p5s* graph. Figs. 7(a) and 7(b) show TCP with the connection order undetermined and enforced, respectively. They reveal two patterns. Firstly, as the *Image* size increases so does the latency. By contrast, Fig. 7(c) shows that LOT is not sensitive to the image size. Secondly, when the connection order is enforced, it becomes apparent that the time it takes each subscriber to receive a given *Image* depends not only on the *Image* size but also how many subscribers have previously established a connection with the publisher. The latter is also observed for TZC in Fig. 7(d) where the connection order was also enforced. We define this behaviour as the late subscribers being treated *unfairly*. We also note that, in spite of the unfairness pattern still being visible, the gap between the median latencies of the most deprived and favoured siblings is much smaller for LOT, e.g. 60 microseconds vs 7 milliseconds for 16MB TCP image payloads and 80 microseconds for TZC at all payload sizes.

We conclude that since LOT, TZC and TCP rely on items being pushed and removed from queues - placed in shared memory and/or kernel space - the unfairness pattern is present in all cases (as expected), but with a negligible latency gap for the LOT protocol.

*C. TCP Buffering Delays Small Payloads*

In order to minimise network traffic at the transport layer, TCP transport may make use of Nagle's algorithm [15] which allows data to be buffered and transmitted together. This effect is shown in Fig. 6(d), where data packets smaller than the TCP buffer size (512B and 1KB in our case) are buffered and sent in bursts, resulting in the latencies being grouped in diamond-like clusters. However, the diamond effect is not observed with LOT, which exhibits steady behaviour and thus fulfills the determinism requirement for real-time systems without requiring Nagle's algorithm to be disabled (i.e. via the TCP_NODELAY socket option). Buffering delays may have an adverse effect on hybrid communication mechanisms such as TZC [2] where one part of the message is transmitted through TCP and the other through shared memory.

*D. Multiple subscribers vs multiple publishers*

Fig. 7(b) and Fig. 5(a) compare the TCP latencies between two symmetric ROS graphs, *1p5s* and *5p1s*. The order of connection establishment has been enforced in both cases to interpret the results more easily. We repeat the analysis of the two symmetric ROS graphs for LOT (Figs 7(c) and 5(b)) and TZC (Figs 7(d) and 5(c)). As expected, the TCP and TZC *1p5s* graph confirms the expected unfairness to late joiners. The *5p1s* case, shown in Fig. 5(a), reveals an interesting fact: all the publishers treat the subscriber fairly, with no obvious periodicity pattern, except for 16MB *Image* size where delays in the order of seconds have been observed. The LOT and TZC counterparts, in Figs 5(b) and 5(c), display much smaller variation in latency of about 50us between publishers. However, the average absolute delay is 225us in the case of LOT, compared to 375us for TZC.

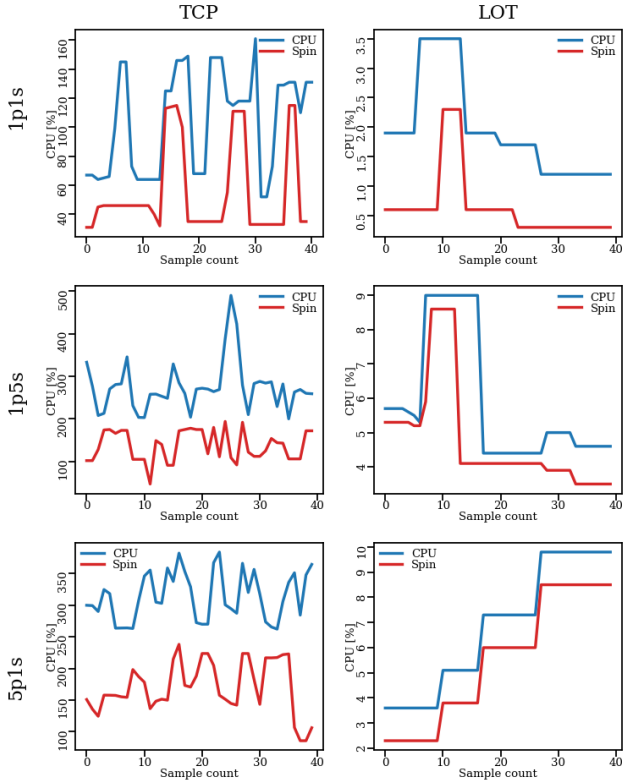

Fig. 8: Activity timeline broken down per protocol and ROS graph demonstrating the utilisation of the twin processors. For clarity only 40 samples have been selected in each plot. The Y axis denotes the utilisation level of the 56 cores. Percentages higher than 100% indicate the utilisation of more than one core. The CPU (blue) line denotes the cumulative utilization of the cores while the code has been executed. This includes the time used for spinning which is also depicted in red. Note the difference in vertical scale between the TCP and LOT cases, and that the CPU utilisation is significantly smaller in the LOT case.

TABLE 1. SUMMARY OF HOTSPOTS

| Case | Spin Time (secs) | Most active functions | CPU time (secs) |
|---|---|---|---|
| TCP *1p1s* | 26.40 | send | 14.10 |
| | | memcpy | 13.44 |
| | | recv | 11.14 |
| LOT *1p1s* | 0.39 | pthread_cond_timedwait | 0.47 |
| | | pthread_mutex_unlock | 0.26 |
| TCP *1p5s* | 78.34 | recv | 45.05 |
| | | memcpy | 43.37 |
| | | send | 32.31 |
| LOT *1p5s* | 2.16 | pthread_cond_timedwait | 1.32 |
| | | pthread_mutex_unlock | 0.51 |
| TCP *5p1s* | 103.15 | send | 56.19 |
| | | memcpy | 48.26 |
| | | recv | 45.67 |
| LOT *5p1s* | 2.63 | pthread_cond_timedwait | 1.44 |
| | | pthread_mutex_unlock | 0.65 |
| | | pthread_cond_broadcast | 0.15 |

Both the LOT and TZC *1p5s* graphs in Fig. 7 show the same subscriber unfairness seen for TCP, but on a vastly smaller scale - tens of microseconds vs tens of milliseconds - with the LOT unfairness visibly smaller than that of TZC. We therefore conclude that the LOT plots show fairness regardless of the *Image* size and ROS graph. In addition to this the latencies are much smaller and so is their variance.

### E. Host vs Same vs Separate Container

Fig. 3 compares the latency incurred by a *1p1s* graph in different execution environments. The TCP plots reveal a slightly bigger median latency of about 1-2 milliseconds when ROS nodes are run in separate Docker containers instead of within the same container. Completely removing the Docker abstraction results in better performance except for 2MB and 4MB *Image* sizes. The LOT plots reveal that the median latency is on average 220 microseconds regardless of the size of the *Image*s as well as the execution environment. Furthermore, in this case the native OS environment is consistently better than the other two, whilst the *same Docker* container case outperforms that of nodes running in *separate Docker* environments.

### F. Pinned vs OS Processor Allocation

On multi-processor machines with a Non-Uniform Memory Access (NUMA) architecture, performance is also influenced by the allocation by the OS of processes to physical processor cores in an attempt to balance the overall load. When the communicating processes happen to be executed on different NUMA nodes, additional hardware mechanisms are required to move data between memory banks [28] resulting not only in increased latency but also unpredictability.

This may be avoided by fixing - *pinning* - processes to physical processors in the same NUMA node. Fig. 6(b) illustrates the effect of OS vs pinned processor allocation for the TCP case. While for payloads up to 32KB the pinned benchmarks outperform the free running ones by a couple of tens of microseconds, for larger payloads the discrepancy generally becomes more pronounced. For the LOT case in Fig 6(c), the effect of pinning the processor allocation is much reduced.

### G. CPU Utilisation

In this section we inspect the code of the above experiments by means of Intel's VTune Profiler [25] in order to verify our understanding of the system behaviour. For brevity, we select the most relevant results to include in this paper and make the full reports available - alongside our source code - via our public repo at github.com/fiveai/ros_comm. We run each experiment for the same duration of 60 seconds and with the same payload of 16MB to ensure a meaningful comparison.

Fig. 8 compares the CPU activity for TCP and LOT transport protocols and demonstrates that the overall activity is far more intense and less effective in the former case, with tens of seconds spent copying *Image*s multiple times. We also measure time spent on network related kernel calls; these are shown in Table 1, which lists the most active functions in the system, as measured by spin time and CPU time. In contrast, in the LOT experiments, CPU activity is greatly reduced, demonstrating that LOT is able to move the same amount of data with far fewer CPU cycles and less contention.

## V. CONCLUSION AND FUTURE WORK

In this work, we have proposed a serialisation-free, zero copy local IPC mechanism implemented using shared pointers stored in a shared memory segment accessed by several ROS nodes. Our work is informed by a careful review of local IPC mechanisms and the key software abstractions that influence their performance. We have evaluated our mechanism with a series of experiments informed by our experience of building a working runtime AV system.

Compared to existing IPC mechanisms within ROS we have demonstrated that our proposed LOT mechanism is the fastest and the least disruptive in terms of determinism for our needs. Coordinated access to the shared memory regions has been accomplished by introducing higher level, shared memory compatible synchronisation primitives with well understood cost and consistent behaviour. In addition to this, we ported the STL's smart pointers semantics to shared memory via *Boost.Interprocess* library and used them consistently to minimise the data copies and facilitate human reasoning.

ROS2 uses the well established Data Distribution Service (DDS) standard [7] for local IPC, though DDS is bypassed for intra-process communication [8]. In future work, we plan to integrate our LOT mechanism with ROS2 and carry out a similar analysis, comparing it with the FastDDS shared memory transport of eProsima [13]. Further attention should also be paid to the case of shared memory management during system shutdown.